%% file: main.tex
\documentclass[conference]{IEEEtran}
\IEEEoverridecommandlockouts
\usepackage{cite}
\usepackage{amsmath,amssymb,amsfonts}
\usepackage{algorithmic}
\usepackage{graphicx}
\usepackage{subfig}
\usepackage{textcomp}
\usepackage[inline]{enumitem}
\usepackage[colorlinks = true,
            linkcolor = blue,
            urlcolor  = blue,
            citecolor = blue,
            anchorcolor = blue]{hyperref} 
\usepackage{xcolor}
\def\BibTeX{{\rm B\kern-.05em{\sc i\kern-.025em b}\kern-.08em
   T\kern-.1667em\lower.7ex\hbox{E}\kern-.125emX}}
  
\newcommand{\R}{\mathbb{R}}
\newcommand\norm[1]{\left\lVert#1\right\rVert}

\DeclareMathOperator{\VR}{VR}
\DeclareMathOperator{\Dgm}{Dgm}
   
\begin{document}

\title{Activation Landscapes as a Topological Summary of Neural Network Performance}

\author{\IEEEauthorblockN{Matthew Wheeler}
\IEEEauthorblockA{\textit{Department of Medicine} \\
\textit{University of Florida}\\
Gainesville, USA \\
mwheeler1@ufl.edu}
\and
\IEEEauthorblockN{Jose Bouza}
\IEEEauthorblockA{\textit{CISE Department} \\
\textit{University of Florida}\\
Gainesville, USA \\
josejbouza@gmail.com}
\and
\IEEEauthorblockN{Peter Bubenik}
\IEEEauthorblockA{\textit{Department of Mathematics} \\
\textit{University of Florida}\\
Gainesville, USA \\
peter.bubenik@ufl.edu}
}
\maketitle

\begin{abstract}
We use topological data analysis (TDA) to study how data transforms as it passes through successive layers of a deep neural network (DNN).  We compute the persistent homology of the activation data for each layer of the network and summarize this information using persistence landscapes.  
The resulting feature map provides both an informative visualization of the network and a kernel for statistical analysis and machine learning.
We observe that the topological complexity often increases with training and that the topological complexity does not decrease with each layer.
\end{abstract}


\input{introduction}

\input{methods}

\input{experiments}
\input{discussion}
\input{conclusion}

\bibliographystyle{plain}
\bibliography{DL}
\end{document}

%% file: introduction.tex
\section{Introduction}

Deep neural networks (DNNs) have become an indispensable tool for handling and analyzing large amounts of data. 
While they have been extremely successful for classifying complex data sets, the way in which they make decisions is often opaque. 
We will use topological data analysis (TDA) to provide a new summary of a DNN which is designed to illuminate aspects of DNN learning and performance.
We focus on how DNNs transform the topology of the input data manifold and how training affects these topological transformations.  
To this end, we use a TDA method, persistence landscapes, to provide a feature map and kernel for the activations of each of the layers in a DNN and we use this method to summarize changes in the DNN during learning.


\subsection{Topological Data Analysis}

Large data sets can exhibit highly complicated topological and geometric structure that is both difficult to summarize and crucial for understanding the data.  
Topological data analysis (TDA) uses methods based on algebraic topology to quantify and study the shape of data.  
One of the primary tools of TDA, \emph{persistent homology}, gives a complete summary of the homology of a one-parameter family of topological spaces~\cite{elz:tPaS,zomorodianCarlsson:computingPH}. 
As an important application, a collection of points in Euclidean space may be used to produce a one-parameter family consisting of unions of balls centered at these points of uniform increasing radius. 
There are a number of efficient implementations for computing persistent homology~\cite{dionysus,perseus,bauer:phat,ripser}
A variant, called \emph{local homology}, has been used to learn the stratification of data~\cite{MR4224154,MR4081170,Robinson:2018,MR3205297}.

\subsection{Deep Neural Networks} 

Deep Neural Networks (DNNs) are overparametrized models with multiple layers and have been shown to be effective for a variety of machine learning tasks, especially for structured data such as images or text~\cite{Goodfellow:2016}.  
Examples of DNN successes include applications in biology~\cite{Dilena2012,Marblestone2016,Abbasi-Asl2018}, chemistry~\cite{Lusci2013}, engineering~\cite{Nie2020}, medicine~\cite{DeFauw2018}, physics~\cite{Baldi2015}, and speech recognition~\cite{Yu2012,Hannun2014,Graves2013speech}.
We will limit our scope to a particular DNN architecture, the multi-layer perceptron (MLP), whose layers consists of an affine transformation followed by a non-linear function, although our method can also be used for other deep learning architectures. 


\subsection{Previous Work}

The synthesis of TDA and DNNs has been shown to be increasingly potent~\cite{Barnes2021ACS,Tauzin2021giottotdaAT}.
Topological summaries have been used in DNNs since the work of Hofer et al.~\cite{Hofer:2017}
Around the same time, persistence landscapes~\cite{bubenik:landscapes} were used as a layer in DNNs for music audio signals~\cite{Liu:2016} and TDA was used in DNNs for the classification of a dynamical systems time series~\cite{Umeda:2017}. 
More recent work has highlighted the use of differentiable topological layers in DNNs~\cite{Carriere:2019,Gabrielsson:2020}. 
For example, persistence landscapes have been used as part of a robust topological layer for DNNs~\cite{Kim2020}.
%
TDA has been used to to measure disentanglement of DNNs \cite{Zhou2021EvaluatingTD} and the use of
TDA to measure and control the topological loss across training has led to state-of-the-art results in training autoencoders~\cite{Hofer2019,Dindin:2019} and GANs~\cite{Wang}.

An important approach to understanding DNNs is 
the development of visualization tools~\cite{Zeiler2014,Yosinski2015,Hohman2019,Selvaraju2020}.
A recent subset of this literature has used the TDA method Mapper~\cite{singh:mapper} to visualize the activations of the layers in a DNN~\cite{BruelGabrielsson2019,Rathore2019}
and to determine the appropriate DNN architecture for certain image data~\cite{Carlsson:2020b}.
Persistent homology has also been used to measure the complexity of DNNs~\cite{rieck2018neural,Guss:2018,Watanabe2020TopologicalMO}.

Naitzat, Zhitnikov and Lim~\cite{Naitzat:2020} used homology and persistent homology to study how the topology of the collection of data from one input class changes as it passes through the layers of a fully trained DNN.
Their analysis indicated that at each layer the topological complexity decreases.
Here we present a kernel-based approach that builds on their work.

%% file: methods.tex
\section{Methods}

\subsection{Persistent Homology}



Given a collection of points $X = \{x_1,\ldots,x_N\} \subset \R^d$ and $r > 0$, the \emph{Vietoris-Rips complex} is the abstract simplicial complex given by
\begin{equation*}
    \VR_r(X)=\left\{\sigma\subset X \mid\, \sigma \neq \emptyset, \, \forall x,y\in\sigma,\, \norm{x -y} \leq r\,\right\}.
\end{equation*}
For $j \geq 0$, the simplicial homology in degree $j$ of this complex with binary coefficients produces a vector space $H_j(\VR_r(X))$.
The dimension of this vector space, called the $j$th \emph{Betti number} gives the number of connected components and the number of holes in $\VR_r(X)$ for $j =0$ and $1$ respectively.

For $r \leq s$, $\VR_r(X) \subseteq \VR_s(X)$ and there is a corresponding linear map $H_j(\VR_r(X)) \to H_j(\VR_s(X))$. 
These vector spaces and linear maps, called a \emph{persistence module}, are completely described by the \emph{persistence diagram} $\Dgm_k(X)$ which consists of ordered pairs $(a,b)$ giving the values of $r$ at which a topological features appear and disappear in the persistence module.



\subsection{Local homology}

We also use a variant of homology called \emph{local homology} to detect local stratified structure. 
The homology of the points $X \subset \R^d$ relative to the exterior of the open unit ball may be computed as follows. All points outside the unit ball are projected to the unit sphere and we take the homology of a modified Vietoris-Rips complex of the resulting points in which pairwise distances between all points on the unit sphere are set to zero.

\subsection{Persistence Landscapes}


Persistence landscapes are a feature map and kernel for persistent homology.
The following construction of persistence landscapes from persistence diagrams is invertible~\cite{bubenik:landscapes}.
For each $(a,b) \in \Dgm_k(X)$, consider the function
    $$f_{(a,b)}(t)=\text{max}(0,\text{min}(a+t,b-t)).$$
The \emph{persistence landscape} is the sequence of functions
    $\lambda=(\lambda_1,\lambda_2,\ldots)$,
where 
    $$\lambda_k(t)=\text{kmax}\left\{f_{(a,b)}(t)\right\}_{(a,b) \in \Dgm_k(X)}.$$
Note that this sequence is decreasing and has at most as many nonzero terms as the number of elements of the persistence diagram.
Persistence landscapes are elements of a Hilbert space with inner product given by 
\begin{equation*}
    \langle \lambda, \rho \rangle = \sum_{k} \int \lambda_k(t) \rho_k(t) \, dt.
\end{equation*}
With this inner product we have a corresponding norm $\norm{\lambda} = \sqrt{\langle \lambda, \lambda \rangle}$ and metric $d(\lambda,\rho) = \norm{\lambda - \rho}$.
We define the \emph{topological complexity} of a persistence landscape to be its norm.




\begin{figure}
\includegraphics[width=.45\textwidth]{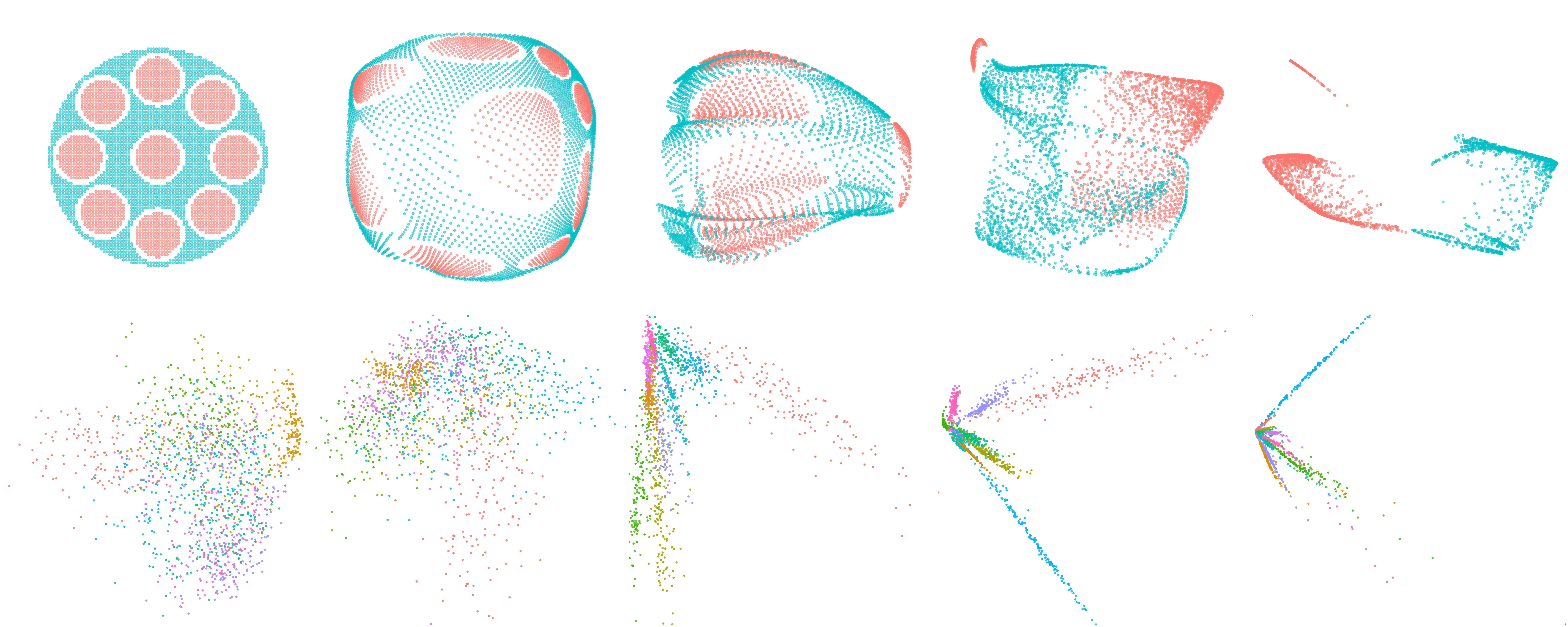}\\
\caption{\textbf{Activations.} 
We consider how input data transforms as it passes through the layers of the network. We visualize these high-dimensional point clouds using the two-dimensional PCA projections.
The top row corresponds to the synthetic data  activations 0, 2, 5, 8, and 11 of a network trained to 99.999\% accurarcy. On the bottom we have the MNIST data activations 0, 1, 3, 5, and 7 of a network trained to 99.5\% accuracy.
Colors indicate ground-truth classes.}

\label{fig:activations}
\end{figure}

\begin{figure*}[ht!]
\centering{
\includegraphics[width=0.57\textwidth]{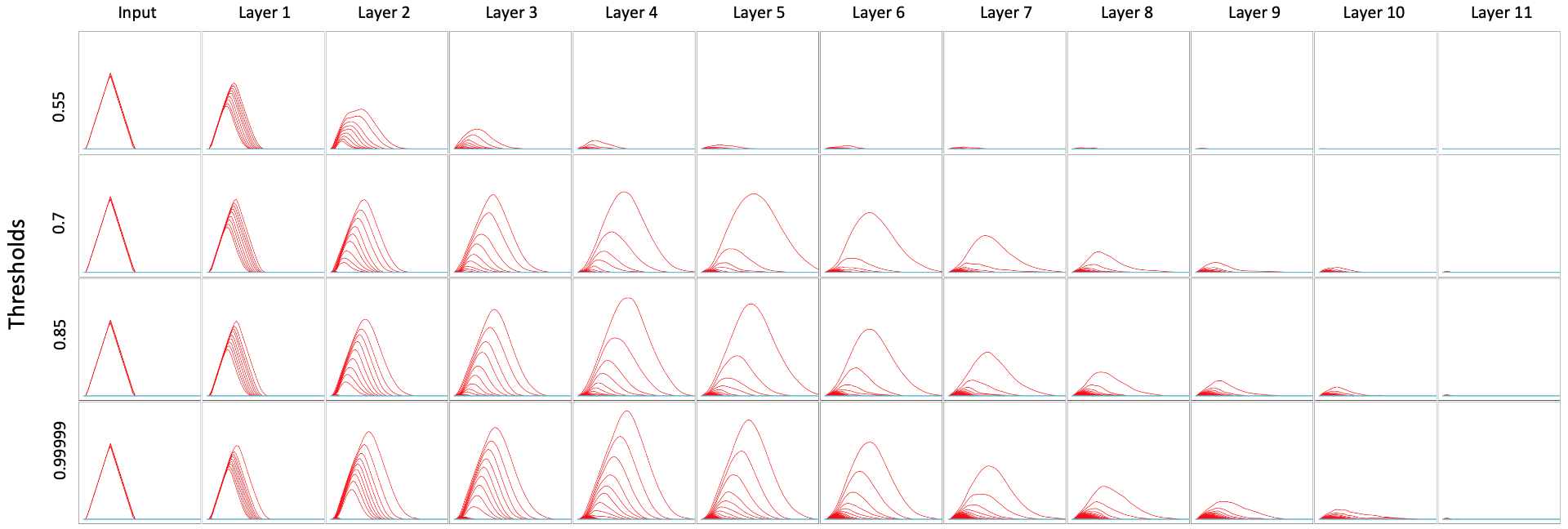}
\includegraphics[width=0.37\textwidth]{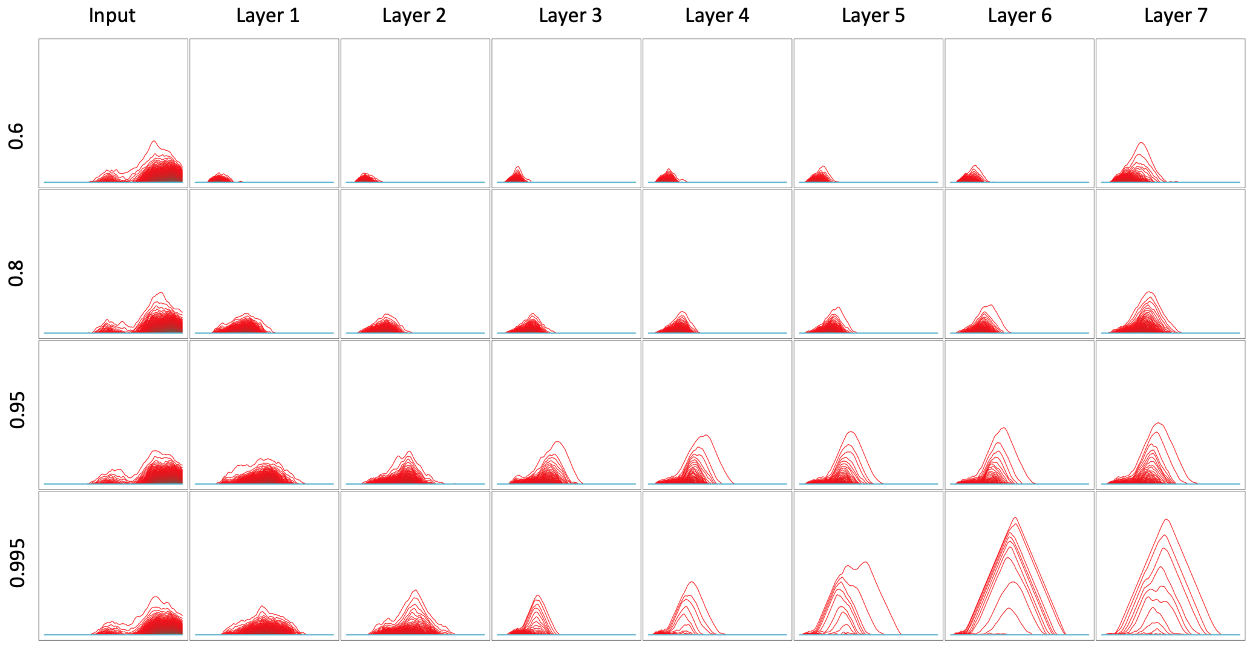}}
\caption{\textbf{Activation landscapes.} 
Left: For our synthetic data and a batch of input data consisting of points from the blue class in Figure~\ref{fig:activations}, we present the activation landscapes for homology in degree one for each layer and various training thresholds averaged across 100 network initializations. Each row represents a training threshold and each column represents a layer of the network.
Right: The activation landscapes for 
MNIST
data with a batch of input data consisting of points from all classes and using local homology in degree one.
} \label{fig:activation-landscapes}
\end{figure*}

\subsection{Activation Landscapes}

We now use persistence landscapes to define a feature map and kernel for multi-layer perceptrons (MLPs).
%
%
Suppose we have an MLP of the form 
\[
    F = P \circ L_N \circ L_{N-1} \circ ... \circ L_1
\]
where the $i$th \emph{layer}, $L_i$, is given by an affine transformation $W_i$ followed by a non-linear function $\sigma$.
For $L_N$, $\sigma$ is omitted.
The network ends with
a softmax function, $P$, mapping the output of $L_N$ to a probability vector. 
We call $N$ the \emph{depth} of the network.

Let $\mathbf{x}$ be a batch of input data to this network. The $i$th \emph{activation} is given by
\[
\mathbf{a}_i =  (L_{i} \circ ... \circ L_1)(\mathbf{x}).
\]
We also let $\mathbf{a}_0 = \mathbf{x}$. 
The \emph{activation landscapes} of network $F$ with input batch $\mathbf{x}$ are computed as follows.
\begin{enumerate}
    \item Center the vectors in $\mathbf{a}_i$ (i.e. translate the mean to the origin) and then normalize so that the maximum distance between the vectors in $\mathbf{a}_i$ is $1$.
    \item Compute the persistence diagram of the Vietoris-Rips complex of the normalized activations for degree $j$.
    \item Compute the corresponding persistence landscape.
\end{enumerate}
We denote the persistence landscapes for layer $i$ by $\lambda^{(i)}$ and call it the $i$th activation landscape. 
Thus to each network and each input we associate a sequence of activation landscapes $\{\lambda^{(0)},\lambda^{(1)},\ldots\lambda^{(N)}\}$.
We may interpret this sequence as a piecewise linear path in Hilbert space which we call  the \emph{activation landscape curve}.  
Alternatively we may consider this sequence as an element of the product Hilbert space, which has inner product 
\begin{equation} \label{eq:activation-curve}
    \langle \lambda, \rho \rangle = \sum_{i=0}^{N} \sum_k \int \lambda_k^{(i)}(t) \rho_k^{(i)}(t) \, dt.
\end{equation}

In the case of local homology, we modify steps (1) and (2) above as follows.
\begin{enumerate}
  \item If the layer does not end with ReLU then center the $\mathbf{a}_i$. Order the resulting vectors by their distance to the origin. 
  Rescale the vectors so the median norm is $1$.
  Replace vectors with norm greater than $1$ with their unit vectors.
  \item Compute the persistence diagram of the Vietoris-Rips complex of a modified distance matrix of these normalized activations in degree $j$ in which all distances between activation vectors whose norms are greater than or equal to the median are set to $0$.
\end{enumerate}



\subsection{Average Activation Landscapes}


Given a fixed network architecture, training data, and batch of input data, the activation landscapes are functions of initial weights from which the networks are trained. If these are randomly determined then the activation landscapes are random variables. The expected values of these random variables may be estimated using the average activation landscapes over a number of samples of initial weights.
Similarly, we may also randomly vary the training data and/or the input data.

%% file: experiments.tex
\section{Experiments}

We now demonstrate that the activation landscapes can illuminate aspects of the training dynamics of a deep neural network. 
%
We ran two experiments to study the activation landscapes of networks trained on synthetic and real data. Both experiments ran on 4 AMD EPYC 7702 Rome 2.0 GHz Cores, 128 GB of system RAM and an RTX 2080 Ti GPU.
%
We provide a highly configurable and tested pipeline for computation of activation landscapes in the \href{https://github.com/jjbouza/tda-nn}{nn-activation-landscapes} Python package. This package integrates with the popular deep learning library PyTorch \cite{paszke2019pytorch}.

\subsection{Synthetic Data}


Following~\cite{Naitzat:2020}, we constructed synthetic data composed of two categories. The first category $C_1$ consists of 9 disks while the second $C_2$ is the complement of the first category within a larger disk. We used an evenly spaced square lattice on the plane to sample points from each category with a combined total of 14,235 points. See the leftmost image in Figure~\ref{fig:activations}. 

We generated 100 trained networks, each with a different weight initialization, for this data set. All network architectures consisted of an MLP with $11$ fully connected layers, the first ten layers of width 15 and the last (output) layer of width 2. Each layer included a ReLU activation except the last layer (which outputs logit vectors). Networks were trained on training data consisting of $90 \%$ of the 14,235 points. Adam~\cite{kingma2014adam} optimizer was used with an initial learning rate of $0.04$. We took snapshots of the weights during the training at prescribed training accuracy thresholds between $50 \%$ and $100 \%$. All networks trained to $100 \%$ training accuracy and had very low generalization error.

To compute activation landscapes we used a batch of input data consisting of a sublattice of $28.52 \%$ of the original data. This sublattice included points from both the testing and training data. We used homology in degree one. Persistence landscapes were discretized with width $0.001$. 

We obtained a three parameter family of activation landscapes across: 
\begin{enumerate*}
    \item networks with different initialization weights and choices of training data;
    \item training thresholds; and
    \item layers of each network.
\end{enumerate*}
For each training threshold and layer, we averaged the activation landscapes across the trained networks. 

\subsection{Real Data}

For our second experiment, we used the MNIST database of $28 \times 28$ pixel images of handwritten digits with $60,000$ training images and $10,000$ test images. 
We generated $10$ trained networks. 
The flattened MNIST images were fed into an MLP with $7$ fully connected layers with feature dimensions $784 \to 128 \to 64 \to 64 \to 64 \to 64 \to 64 \to 10$. 
As in the synthetic experiment, each layer was followed by a ReLU activation function except for the last. 
We again used an Adam optimizer for training with an initial learning rate of $0.04$. 
We took snapshots of the weights at various training accuracy thresholds between $20 \%$ and $100 \%$.
All networks trained to near $100 \%$ training and test accuracy.
We computed activation landscapes on a randomly sampled subset of size $2,000$ from the union of the training and test images (approximately $2.1 \%$ of the images).
We used local homology in degree one and  persistence landscapes were discretized to width $0.005$. 
We obtained a four parameter family of activation landscapes across: 
\begin{enumerate*}
    \item networks with different initialization weights and choices of training data;
    \item training thresholds; 
    \item layers of each network; and
    \item subsamples of input data.
\end{enumerate*}
For each training threshold and layer, 
activation landscapes were averaged over 
choices of trained networks and input data.

\subsection{Results}

For a choice of training data and a fixed network architecture, we consider the activations and activation landscapes for various network initializations, thresholds, and layers. 
In Figure~\ref{fig:activations} we visualize the activations for some of the layers in a particular fully trained network for both our synthetic data and real data.
In Figure~\ref{fig:activation-landscapes}, we plot the average activation landscapes for all layers and a selection of training accuracy thresholds for homology in degree 1 for $C_2$ for our synthetic data on the left and for local homology in degree 1 for our real data on the right.
For the synthetic data, observe that the nine holes in the blue input data in Figure~\ref{fig:activations} are detected by the activation landscapes of the starting layers in Figure~\ref{fig:activation-landscapes}. 
For our real data, observe in Figure~\ref{fig:activations} that inputs of a given class as they pass through the network seem to cluster on one-dimensional subspaces emanating from the origin. These features are detected by the activation landscapes in ending layers of Figure~\ref{fig:activation-landscapes}.

\begin{figure}
\centering
\includegraphics[width=.24\textwidth]{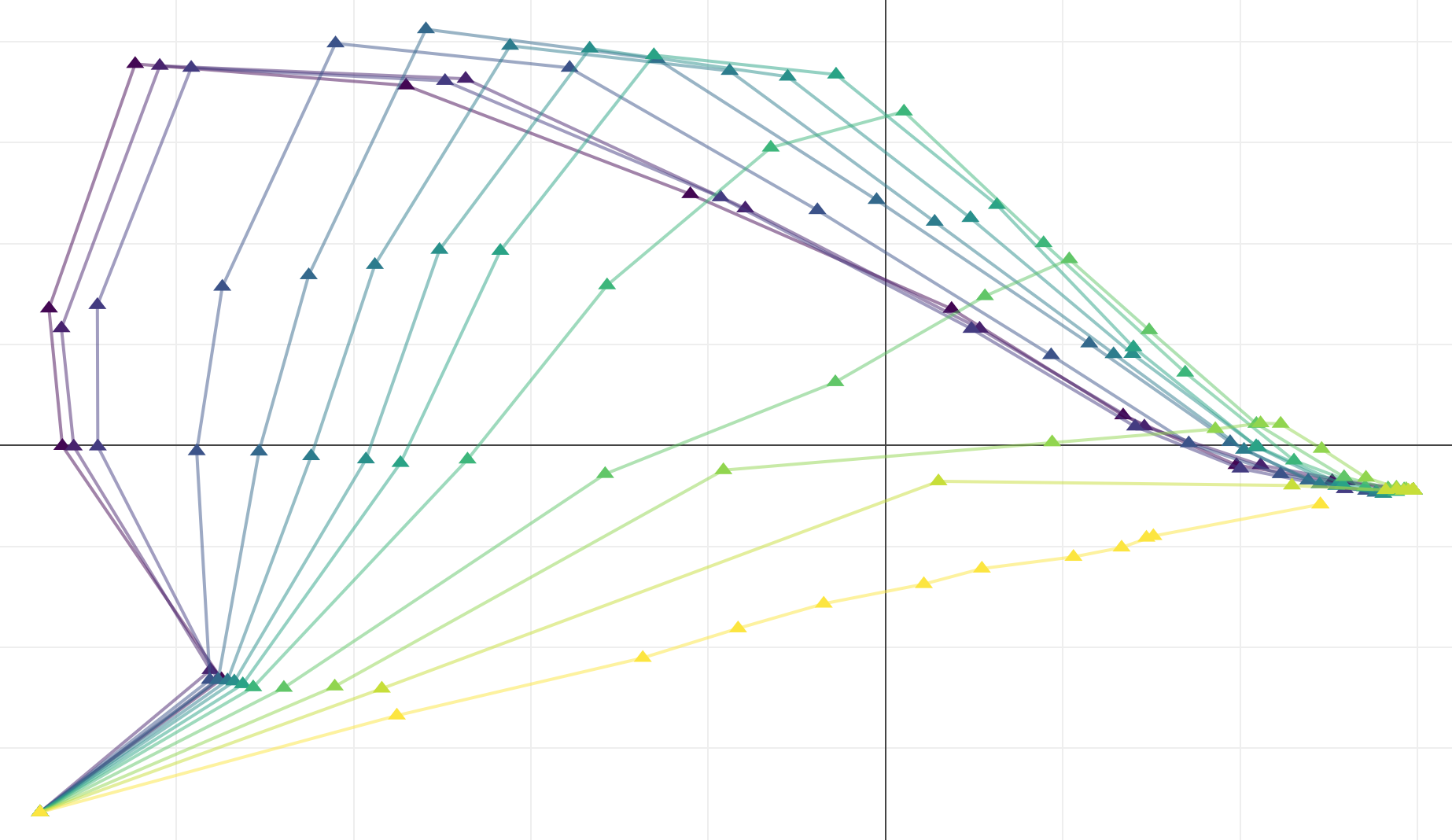}
\includegraphics[width=.24\textwidth]{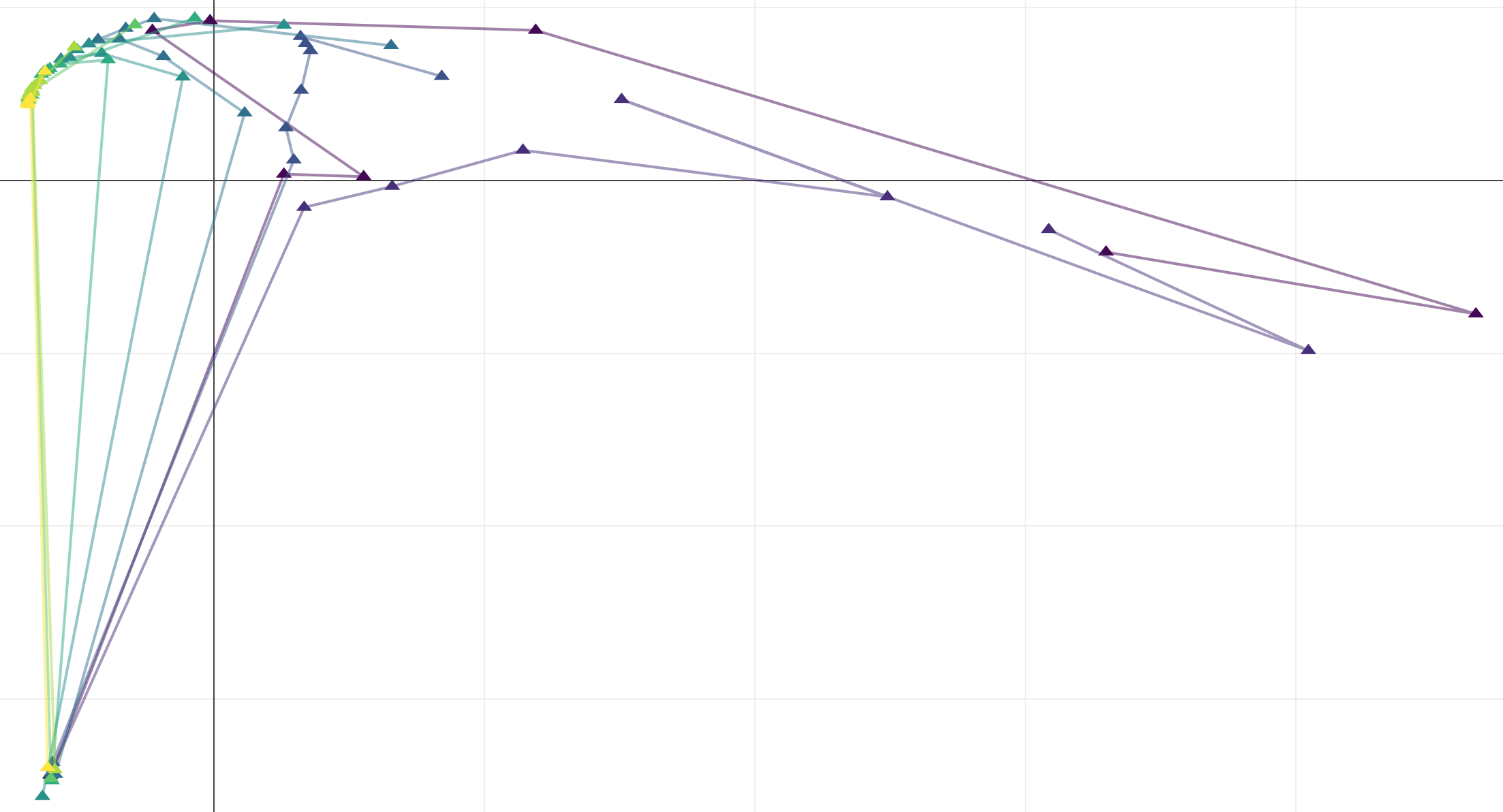}
\caption{\textbf{Activation curves.} Two-dimensional PCA projections of the activation landscapes from Figure~\ref{fig:activation-landscapes} using a larger number of training accuracy thresholds, with consecutive layers connected by line segments (increasing layers generally from left to right) and increasing accuracy thresholds colored from yellow to purple. }
\label{fig:activation-curves}
\end{figure}

In Figure~\ref{fig:activation-curves} we give a 2D projection of the activation landscapes for all of the layers and a range of learning accuracy thresholds including the four in Figure~\ref{fig:activation-landscapes} for both our synthetic and real data.
This is a 2D projection of a sequence of piecewise linear high-dimensional curves.

\begin{figure}
\centering
\includegraphics[width=.24\textwidth]{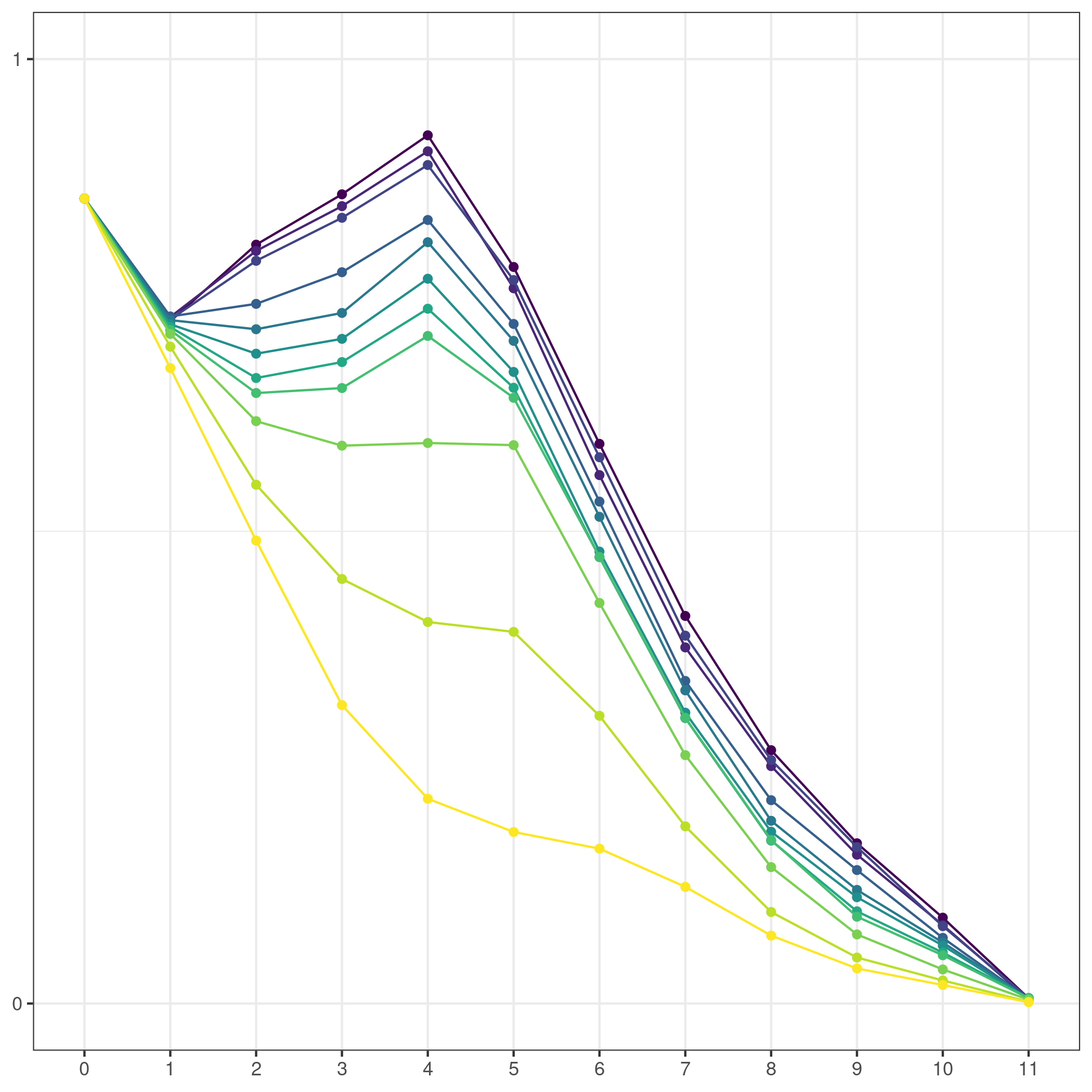}
\includegraphics[width=.24\textwidth]{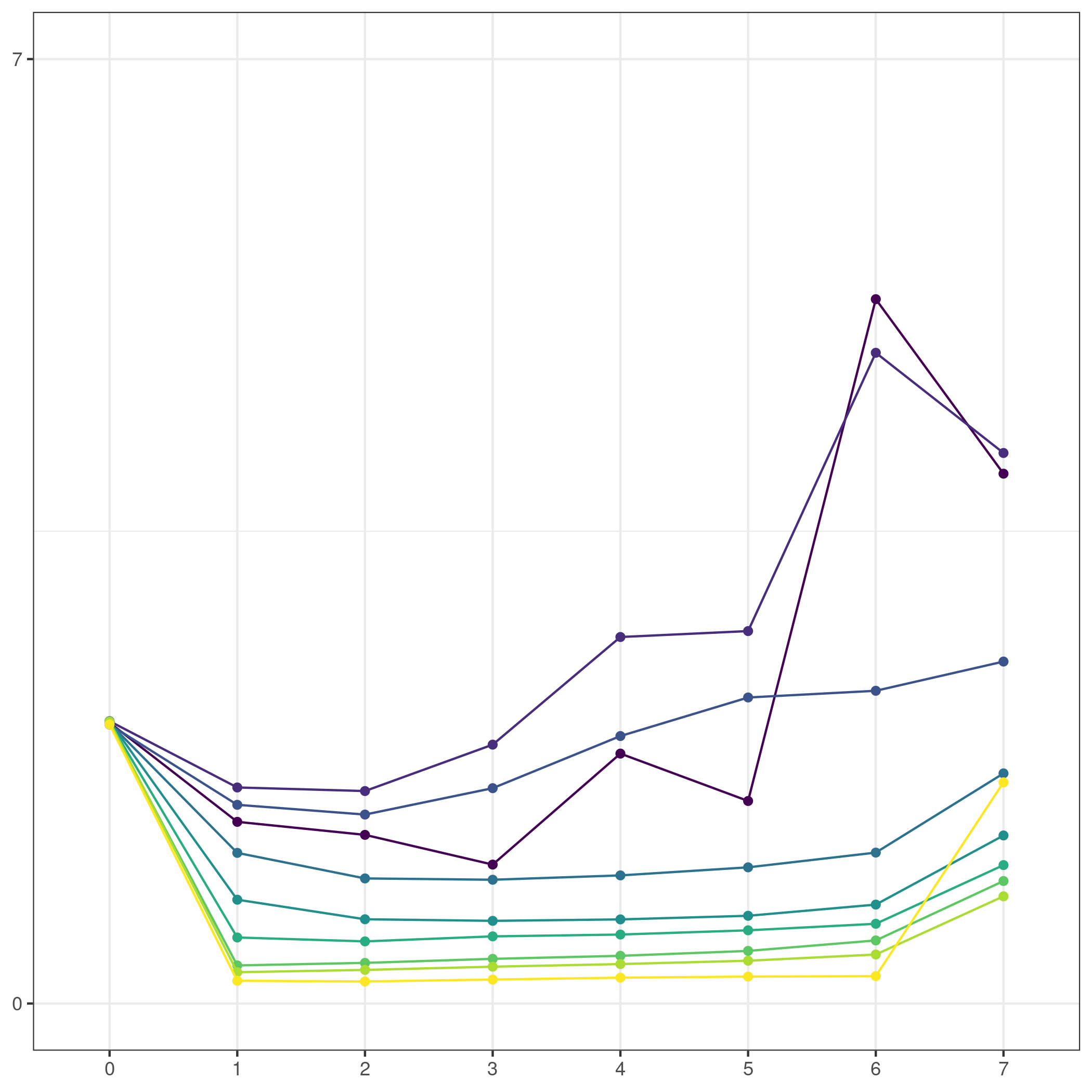}
\caption{\textbf{Topological complexity.} Average norms 
of the activation landscapes 
from Figure~\ref{fig:activation-landscapes} using a larger number of training accuracy thresholds
with consecutive layers connected by line segments and increasing accuracy thresholds colored from yellow to purple.} 

\label{fig:topological-complexity}
\end{figure}

In Figure~\ref{fig:topological-complexity} we give the norms of the activation landscapes of all of the layers and a range of learning accuracy thresholds including the four in Figure~\ref{fig:activation-landscapes}.
Observe that these curves are not monotonically decreasing and that this effect increases as the learning accuracy threshold increases.

Finally we test if the activation landscapes for the various training accuracy thresholds whose averages are shown in Figure~\ref{fig:activation-landscapes} are statistically significantly different. We perform permutation test with $N=100,000$ on the test statistic given by the distance between the average activation landscapes using the inner product in \eqref{eq:activation-curve}. Table~\ref{table:permutation_test} shows the p values. 
In all cases, the difference is statistically significant for the significance level of 0.05. 

\begin{table}
\centering
\caption{The p-values for differences between the activation landscapes in Figure~\ref{fig:activation-landscapes} between the various training thresholds. Top: Synthetic data. Bottom: MNIST data.}
\begin{tabular}{ |c| c c c c| } 
 \hline
  \textbf{Training Accuracy} & $55 \%$ &  $70 \%$ & $85 \%$ & $99.999 \%$ \\ 
 \hline
 $55 \%$ & $\cdot$ & $\cdot$ & $\cdot$ & $\cdot$ \\
 $70 \%$ & $0.01434$ & $\cdot$ & $\cdot$ & $\cdot$\\ 
 $85 \%$ & $0.00000$ & $0.00792$ & $\cdot$ & $\cdot$\\ 
 $99.999 \%$ &$0.00000$ & $0.00000$ & $0.01579$ & $\cdot$ \\ 
 \hline
  \textbf{Training Accuracy} & $60 \%$ &  $80 \%$ & $95 \%$ & $99.5 \%$ \\ 
 \hline
 $55 \%$ & $\cdot$ & $\cdot$ & $\cdot$ & $\cdot$ \\
 $70 \%$ & $0.00000$ & $\cdot$ & $\cdot$ & $\cdot$\\ 
 $85 \%$ & $0.00000$ & $0.00000$ & $\cdot$ & $\cdot$\\ 
 $99.5 \%$ &$0.00000$ & $0.00000$ & $0.00000$ & $\cdot$ \\ 
 \hline
\end{tabular}
\label{table:permutation_test}
\end{table}

%% file: discussion.tex
\section{Disussion}

In Figures \ref{fig:activation-landscapes} and \ref{fig:activation-curves} we observe that as the training accuracy threshold increases the activation landscapes are converging toward the expected activation landscape of a fully trained network. 
Furthermore, the latter seems to accentuate the most significant topological features of the activations.
In Figure~\ref{fig:topological-complexity} we plot the norms of these activation landscapes. 
We make two striking observations.
First, the topological complexity of the activations does not decrease monotonically as the data passes through the layers of the well trained networks, contradicting a previous observation on the topology of deep neural networks~\cite{Naitzat:2020}.
Second, as the training threshold increases the topological complexity of the activations tends to increases.

%% file: conclusion.tex
\section{Conclusion}

Our activation landscapes provide a powerful and innovative way for analyzing the evolution of topological information as it passes through the layers of a deep neural network
and how this changes throughout the training process.
Our method has a number of benefits over previous instances.  
First, it provides a complete summary of the persistent homology of the activations in each layer in the network.
Second, as a feature map and kernel it allows us to easily apply statistics and machine learning in our analysis.
